\documentclass[conference]{IEEEtran}
\IEEEoverridecommandlockouts

\usepackage{tikz}
\usepackage{amssymb}
\usepackage{float}
\usepackage{mathrsfs}
\usepackage{tabularx}
\usepackage{booktabs}
\usepackage{longtable}
\usepackage{subcaption}
\usepackage{caption}
\usepackage{amsmath}
\usepackage{todonotes}
\usepackage{xparse} 

\usepackage{svg}
\usepackage{subcaption}
\usepackage{graphicx}
\usepackage{booktabs}

\usepackage{stfloats}

\setsvg{inkscapelatex=false} 

\usetikzlibrary{arrows.meta,positioning,fit}
\usetikzlibrary{fit,shapes.geometric,positioning, calc}
\tikzset{
  nodeStyle/.style={draw, rounded corners=2pt, inner sep=1pt, minimum width=6mm, minimum height=4mm, font=\scriptsize},
  edgeStyle/.style={-Latex},
  gsn-goal/.style={draw, rectangle, minimum width=1cm, minimum height=0.8cm, text width=1cm, align=center, font=\scriptsize},
  gsn-solution/.style={draw, circle, minimum width=1.0cm, text width=1.0cm, align=center, font=\scriptsize},
  gsn-strategy/.style={draw, trapezium, trapezium left angle=70, trapezium right angle=110, minimum width=1.8cm, minimum height=0.8cm, text width=1.5cm, align=center, font=\scriptsize},
  gsn-assumption/.style={draw, ellipse, minimum width=1cm, minimum height=0.5cm, text width=1cm, align=center, font=\scriptsize},
  gsn-context/.style={draw, rectangle, rounded corners=8pt, minimum width=1.6cm, minimum height=0.5cm, text width=1.8cm, align=center, font=\scriptsize},
  gsn-supported/.style={-Latex},
  gsn-incontext/.style={-{Latex[open]}},
  gsn-fusion/.style={draw, circle, inner sep=1pt, minimum size=7pt, font=\tiny}
}

\newcommand{\tofilled}{%
  \mathrel{\tikz[baseline=-0.5ex]\draw[-{Stealth[length=1ex,width=1ex]},line width=0.35pt]
    (0,0)--(1.3em,0);}}
\newcommand{\toopen}{%
  \mathrel{\tikz[baseline=-0.5ex]\draw[-{Triangle[open,length=1ex,width=1ex]},line width=0.35pt]
    (0,0)--(1.3em,0);}}

\usepackage{booktabs,array,longtable}
\usepackage{tikz}
\usetikzlibrary{positioning,arrows.meta}

\newtheorem{definition}{Definition}

\def\BibTeX{{\rm B\kern-.05em{\sc i\kern-.025em b}\kern-.08em
    T\kern-.1667em\lower.7ex\hbox{E}\kern-.125emX}}

\newcommand\copyrighttext{%
  \footnotesize \textcopyright 2026 IEEE. Personal use of this material is permitted.
  Permission from IEEE must be obtained for all other uses, in any current or future
  media, including reprinting/republishing this material for advertising or promotional
  purposes, creating new collective works, for resale or redistribution to servers or
  lists, or reuse of any copyrighted component of this work in other works.}
\newcommand\copyrightnotice{%
\begin{tikzpicture}[remember picture,overlay]
\node[anchor=south,yshift=10pt] at (current page.south) {\fbox{\parbox{\dimexpr\textwidth-\fboxsep-\fboxrule\relax}{\copyrighttext}}};
\end{tikzpicture}%
}

\begin{document}
\newcommand{\dbra}[1]{\ensuremath{\llbracket #1 \rrbracket}} 

\NewDocumentCommand{\opinion}{m o}{%
  \ensuremath{\omega_{#1}\IfValueT{#2}{^{#2}}}%
}
\NewDocumentCommand{\binop}{m}{\ensuremath{\langle b_{#1},\, d_{#1},\, u_{#1},\, a_{#1}\rangle}} 

\newcommand{\belief}[1]{\ensuremath{b_{#1}}}
\newcommand{\disbelief}[1]{\ensuremath{d_{#1}}}
\newcommand{\uncert}[1]{\ensuremath{u_{#1}}}
\newcommand{\baserate}[1]{\ensuremath{a_{#1}}}

\newcommand{\vbelief}[1]{\ensuremath{\mathbf{b}_{#1}}}
\newcommand{\vbaserate}[1]{\ensuremath{\mathbf{a}_{#1}}}
\newcommand{\uncertV}[1]{\ensuremath{u_{#1}}}

\newcommand{\marginalop}[1]{\dbra{#1}}

\newcommand{\deduces}{\mathrel{\Vert}} 
\newcommand{\deducedopinion}[2]{\opinion{#1\deduces #2}}

\newcommand{\given}{\mathrel{\mid}}   
\newcommand{\conditionalopinion}[2]{\opinion{#1\given #2}}
\NewDocumentCommand{\conditionals}{s m m}{%
\ensuremath{
  \left( \omega_{#2 \mid #3},\, \omega_{#2 \mid \lnot\, \IfBooleanTF{#1}{(#3)}{#3}} \right)%
}
}

\newcommand{\proj}[1]{\ensuremath{p_{#1}}}
\newcommand{\projof}[1]{\ensuremath{\operatorname{proj}\!\left(#1\right)}}
\newcommand{\expect}[1]{\ensuremath{\mathbb{E}\!\left[#1\right]}}
\newcommand{\expecop}[1]{\ensuremath{\mathbb{E}\!\left[\opinion{#1}\right]}}

\newcommand{\projrule}[1]{\ensuremath{\belief{#1} + \baserate{#1}\,\uncert{#1}}} 
\newcommand{\normop}[1]{\ensuremath{\belief{#1} + \disbelief{#1} + \uncert{#1} = 1}}

\newcommand{\evidence}[1]{\ensuremath{\mathbf{r}_{#1}}}
\newcommand{\evidp}[1]{\ensuremath{r_{#1}^{+}}}
\newcommand{\evidm}[1]{\ensuremath{r_{#1}^{-}}}
\newcommand{\dirichlet}[1]{\ensuremath{\operatorname{Dir}\!\left(#1\right)}}
\newcommand{\BetaDist}[1]{\ensuremath{\operatorname{Beta}\!\left(#1\right)}}
\newcommand{\alphaDir}[1]{\ensuremath{\boldsymbol{\alpha}_{#1}}}

\newcommand{\cumfusion}{\ensuremath{\mathbin{\oplus}}}        
\newcommand{\afusion}{\ensuremath{\mathbin{\underline{\oplus}}}} 
\newcommand{\constraintfusion}{\ensuremath{\mathbin{\odot}}} 
\newcommand{\discount}{\ensuremath{\mathbin{\otimes}}}        
\newcommand{\deduction}{\ensuremath{\mathbin{\circledcirc}}}  
\newcommand{\abduction}{\ensuremath{\tilde{\mathbin{\circledcirc}}}} 
\newcommand{\inversion}{\ensuremath{\tilde{\phi}}}                
\newcommand{\weightedfusion}{\ensuremath{\mathbin{\hat{\oplus}}}} 
\newcommand{\ccfusion}{\ensuremath{\mathbin{\ooalign{%
  \hfil \raisebox{0.1ex}{\tiny CC}\hfil\crcr
  $\bigcirc$}}}}  
\newcommand{\unfusion}{\ensuremath{\mathbin{\ominus}}}        

\newcommand{\negate}[1]{\ensuremath{\overline{#1}}}
\newcommand{\co}[1]{\ensuremath{#1^{\complement}}}
\newcommand{\domain}[1]{\ensuremath{\mathbb{#1}}}
\newcommand{\variable}[1]{\ensuremath{\mathcal{#1}}}
\title{Towards a compositional semantics for quantitative confidence assessment in assurance arguments}
\author{\IEEEauthorblockN{Benjamin Herd}
\IEEEauthorblockA{\textit{Fraunhofer IKS}\\
Munich, Germany \\
0000-0001-6439-8845}
\and
\IEEEauthorblockN{Jessica Kelly}
\IEEEauthorblockA{\textit{Fraunhofer IKS}\\
Munich, Germany\\
0009-0003-0508-2367}
\and
\IEEEauthorblockN{Jan Sabsch}
\IEEEauthorblockA{\textit{Luxoft GmbH}\\
Munich, Germany \\
jan.sabsch@dxc.com}
\and
\IEEEauthorblockN{Lydia Gauerhof}
\IEEEauthorblockA{\textit{Robert Bosch GmbH} \\
Renningen, Germany \\
0000-0002-3504-0040}
}

\maketitle
\copyrightnotice
\begin{abstract}
Assurance arguments provide a clear and structured way to explain why stakeholders should trust that a system satisfies certain properties, yet widely used notations, e.g.\ Goal Structuring Notation (GSN), typically lack an operational semantics for deriving \emph{assurance confidence}. Existing approaches address structure and soundness but largely reason over truth values, not over confidence in the justification of claims. Subjective Logic (SL) offers a calculus of belief, disbelief, and uncertainty with operators for combining opinions, enabling confidence propagation under incomplete, conflicting, or subjective evidence. However, existing SL-based approaches do not provide a uniform, compositional semantics that covers all argument elements and relations to enable overall confidence assessment. We propose a confidence semantics that represents argument elements as SL opinions and maps relations between elements to SL operators modelling how confidence flows, effectively turning the argument into an analyzable confidence network. The approach provides explicit warrants, principled handling of context, preserved provenance, and compatibility with GSN, along with practical guidance using an exemplary assurance confidence assessment.
\end{abstract}

\section{Introduction}
Assurance arguments provide a clear and structured means to communicate why stakeholders should believe that a system satisfies certain properties, e.g.\ safety. While notations such as the Goal Structuring Notation (GSN) \cite{kelly2004goal} express the \textit{structure} of reasoning, they generally lack an operational semantics that defines how \textit{assurance confidence} -- i.e.\ trust in the validity of the argument itself -- can be formally obtained. Proposed formalisms ranging from logical interpretations of GSN \cite{Bandur2015} to compositional safety case patterns \cite{denney2016composition} address aspects of structure or soundness but typically reason over truth values of claims, not over confidence in their justification.  

Prior work has leveraged Subjective Logic (SL), a variant of probabilistic logic that models confidence as \textit{subjective opinions}, i.e.\ tuples of belief, disbelief, and uncertainty, for assurance confidence assessments \cite{Duan2015,YuanSubjectiveLogic2017,HerdDeductiveApproach}. SL provides a wide range of algebraic operators to combine such opinions and express the argument's inferential structure as a \textit{network of confidence propagation}. This allows for automated confidence assessment also in cases where quantitative measurements are incomplete or where uncertainty must be carried forward, e.g.\ in the case of conflicting, subjective, or partial evidence. However, existing approaches do not yet provide a general-purpose, compositional operational semantics that addresses all elements across an entire argument graph and enables overall confidence assessment.

We build upon this prior work by providing a uniform, graph-wide, compositional semantics where each argument element (e.g., goal, evidence, assumption) is represented as an opinion in SL (which encapsulates its current confidence state) and every relation between elements is interpreted as a corresponding SL operator that governs how confidence flows between elements. The resulting semantics is both formal and compositional and yields a directly analyzable confidence network that mirrors the argument's structure. The proposed semantics equips the assurance arguments with a precise quantitative backbone where confidence can be computed systematically. This bridges the gap between qualitative argumentation and quantitative confidence assessment and enables automation and modular reasoning without departing from familiar argument notations. We make the following contributions: 

\begin{enumerate}
    \item We propose a compositional semantics for the computation of assurance argument confidence. This approach:
    \begin{itemize}
        \item represents confidence in argument elements as SL opinions and models confidence propagation between the elements of an argument end-to-end while preserving modularity and provenance;
        \item supports the representation of different reasoning strategies and argument patterns, including conditional treatment of contextual information;
        \item considers `second-order confidence' in the argument by modeling the validity of each inferential step as an explicit warrant that can be justified; 
    \end{itemize}
    \item We provide practical guidance on deriving assurance confidence and illustrate the approach using a simple exemplary confidence assessment. 
\end{enumerate}

The paper is structured as follows. Section \ref{sec:background} provides background on safety assurance and SL. Section \ref{sec:mapping} presents the formal mapping of relations between argument elements to SL operators. Section \ref{sec:argument_patterns} develops modular composition patterns, integrates contextual assumptions, and provides guidance on the elicitation of justifications for inferential steps. Section \ref{sec:example} walks through an overall confidence assessment. Related work is covered in Section \ref{sec:related_work}, and Section \ref{sec:discussion} contains a discussion of challenges and concluding remarks. 

\section{Background}\label{sec:background}

\subsection{Safety assurance}
\textit{Assurance} is defined as grounds for justified confidence that a \textit{claim} has been or will be achieved \cite{IEEE_15026}. A \textit{claim} is defined as a true-false statement about the limitations on the values of the claim's property and limitations on the uncertainty of the property’s values falling within these limitations. \cite{IEEE_15026} also defines an \textit{assurance argument} as a reasoned, auditable artefact that supports the contention that its top-level claim is satisfied, including systematic arguments and its underlying evidence and explicit assumptions that support the claim(s). As such, the assurance argument communicates the relationship between evidence and the safety objectives. 

\begin{figure}[bt]
\begin{tikzpicture}[node distance=0.8cm and 1.0cm, every node/.style={transform shape}]

\node[gsn-goal, text width=2.3cm] (G1) {G1\\ System is acceptably safe to operate};
\node[gsn-strategy, below=0.6cm of G1, text width=2.5cm] (S1) {S1\\Argument over all hazards};
\node[gsn-assumption, text width = 1.8cm, left=0.8cm of G1] (A1) {A1\\All hazards are completely and correctly identified};

\node[gsn-goal, below left=0.5cm and 0.4cm of S1, text width=2cm] (G2) {G2\\Hazard 1 mitigated};
\node[gsn-goal, below right=0.5cm and 0.4cm of S1, text width=2cm] (G3) {G3\\Hazard 2 mitigated};

\node[gsn-strategy, below=0.5cm of G3, text width=2.5cm] (S3) {S3\\Argument by alternative mitigations};

\node[gsn-goal, below=0.35cm of S3, text width=1.6cm, xshift=-1cm] (G6) {G6\\Mitigation A implemented};
\node[gsn-goal, below=0.35cm of S3, text width=1.6cm, xshift=1cm] (G7) {G7\\Mitigation B implemented};

\draw[gsn-supported] (G3) -- (S3);
\draw[gsn-supported] (S3.south) -- (G6.north);
\draw[gsn-supported] (S3.south) -- (G7.north);

\node[gsn-strategy, below=0.5cm of G2, text width=2.5cm] (S2) {S2\\Argument over diverse evidence};

\node[gsn-goal, below=0.4cm of S2, xshift=-1cm, text width=1cm] (G4) {G4\\Evidence claim 1};
\node[gsn-goal, below=0.4cm of S2, xshift=1cm, text width=1cm] (G5) {G5\\Evidence claim 2};
\node[diamond, draw, inner sep=2pt, minimum size=7pt, yshift=-0.12cm] (U1) at (G5.south) { };

\node[gsn-solution, below=0.4cm of G4] (Sn1) {Sn1};
\node[gsn-solution, below=0.4cm of G6] (Sn2) {Sn2};
\node[gsn-solution, below=0.4cm of G7] (Sn3) {Sn3};

\draw[gsn-supported] (G1) -- (S1);
\draw[gsn-incontext] (G1) -- (A1);

\draw[gsn-supported] (S1.south) -- (G2.north);
\draw[gsn-supported] (S1.south) -- (G3.north);

\draw[gsn-supported] (G2) -- (S2);

\draw[gsn-supported] (S2.south) -- (G4.north);
\draw[gsn-supported] (S2.south) -- (G5.north);

\draw[gsn-supported] (G4) -- (Sn1);
\draw[gsn-supported] (G6) -- (Sn2);
\draw[gsn-supported] (G7) -- (Sn3);

\node[draw, densely dotted, rounded corners, inner sep=3pt,
      fit=(S1)(G2)(G3)] (boxI) {};
\node[font=\scriptsize, anchor=north west] at (boxI.north west) {I};

\node[draw, densely dotted, rounded corners, inner sep=4pt,
      fit=(S2)(G4)(G5)] (boxII) {};
\node[font=\scriptsize, anchor=north west] at (boxII.north west) {II};

\node[draw, densely dotted, rounded corners, inner sep=4pt,
      fit=(S3)(G6)(G7)] (boxIII) {};
\node[font=\scriptsize, anchor=north west] at (boxIII.north west) {III};
\end{tikzpicture}
\caption{Simple GSN argument representing a composition of (I) conjunctive, (II) fusion, and (III) disjunctive arguments. Adapted from \cite{Hawkins2011}.}
\label{fig:gsn-running-example}
\end{figure}
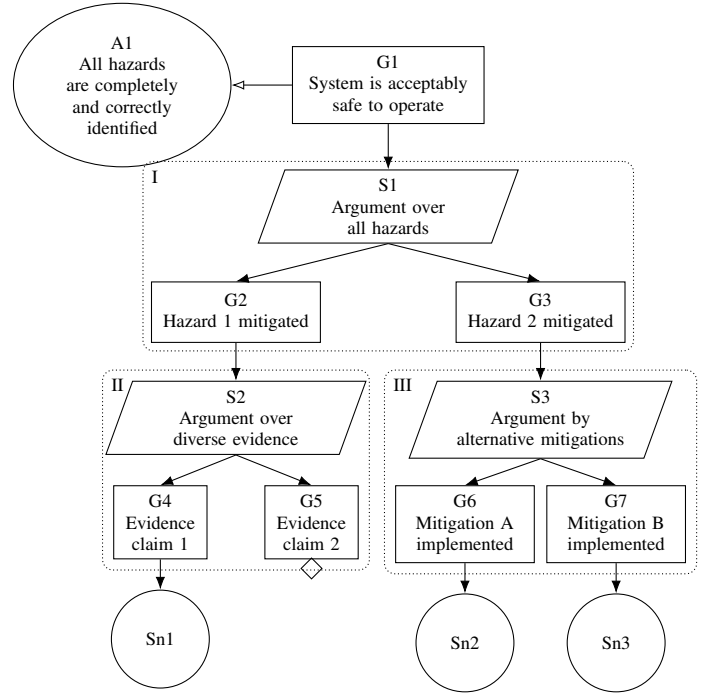

A model-based graphical representation of the argument can aid its communication and evaluation. In this paper we make use of the Goal Structuring Notation (GSN) \cite{kelly2004goal} to illustrate our approach. A simple GSN argument is shown in Fig.~\ref{fig:gsn-running-example}. In GSN, a \emph{goal} is a claim to be justified, a \emph{strategy} explains how a goal is decomposed into sub-goals (i.e., the form of inference), a \emph{solution} is evidence that directly supports a goal (e.g.\ analysis or test results), and an \emph{assumption} is a stated condition taken as true for the argument. Relations capture how these elements connect: \emph{supportedBy} ($\tofilled$) denotes inferential support from strategies, sub-goals, or evidence to a parent goal, while \emph{inContextOf} ($\toopen$) attaches contextual items (assumptions, context, justifications) that frame the interpretation of the connected element.

While notations like GSN provide a clear graphical syntax and naming of argument elements, its semantics remain largely informal. As a result, key aspects are left to expert judgment: the logical reading of a decomposition (e.g., conjunctive vs.\ disjunctive support), the scope and effect of context on claims, the treatment of uncertainty and conflicting evidence, and the aggregation of multiple supports (including their dependence).

\subsection{Subjective Logic}\label{sec:subjective_logic}

Subjective Logic (SL) \cite{josang2016subjective} is a framework for reasoning with uncertain beliefs that combines ideas from probabilistic logic and evidence theory. In contrast to Bayesian or purely logical representations, SL allows for \textit{explicit} reasoning with incomplete or conflicting information and subjectivity. 

The atomic building blocks of SL are \emph{subjective opinions}, and SL offers a range of combination operators that allow for algebraic reasoning. Subjective opinions express beliefs about the truth of propositions under degrees of uncertainty. We form opinions about variables which take their values from \textit{domains} -- defined as mutually exclusive and collectively exhaustive sets of states, events, outcomes, hypotheses, or propositions. Throughout this paper, we consider opinions defined over binary domains $\mathbb{X}=\{x,\bar{x}\}$ corresponding to the two truth states of propositions about assurance argument elements: the proposition that an element is satisfied and the proposition that it is not satisfied. Such opinions are called \emph{binomial} opinions. Two opinions are considered \textit{independent} if their underlying sources of evidence have no overlap -- that is, they are based on distinct bodies of information. This notion of evidential independence differs from statistical independence in probability theory, which refers to the independence of random variables rather than of information sources.

\begin{definition}[Binomial opinion]\label{def:binop}
Let $\mathbb{X} = \{x,\bar{x}\}$ be a binary domain. A binomial opinion about the truth of $x$ is a tuple $\omega_x = (b_x,d_x,u_x,a_x)$ where 
\begin{itemize}
    \item $b_x$ (belief): the belief mass in support of $x$ being \textbf{\textit{true}}
    \item $d_x$ (disbelief): the belief mass in support of $x$ being \textbf{\textit{false}}
    \item $u_x$ (uncertainty): the uncommitted belief mass
    \item $a_x$ (base rate): the \textit{a priori} probability in the absence of committed belief mass (often set to 0.5 for binary domains) 
\end{itemize}
where $b_x,d_x,u_x,a_x \in [0,1]$ and $b_x + d_x + u_x = 1$. 
\end{definition}
\vspace{0.1cm}

\paragraph{Constructing opinions} Given positive evidence $r$ for a claim, negative evidence $s$ and a \textit{non-informative prior weight}\footnote{$W$ ensures that when evidence begins to accumulate (i.e.\ $r$ gets larger), uncertainty $u_x$ decreases accordingly. $W$ is typically set to the same value as the cardinality of the domain (2 in our binary case), thus artificially adding one success $r$ and one failure $s$. Higher values of $W$ require more evidence for uncertainty to decrease.} $W$, a Binomial opinion can be computed as follows: 
\begin{align}
    b_x &= r/(r+s+W)\label{eq:belief}\\
    d_x &= s/(r+s+W)\label{eq:disbelief}\\
    u_x &= W/(r+s+W)\label{eq:uncertainty}
\end{align}
with base rate $a\in[0,1]$. The prior weight $W$ controls how fast uncertainty $u$ decreases as evidence accumulates (commonly $W=2$ in binary domains). Binomial opinions correspond to Beta distributions: with $(r,s)$, $a$, and $W$, the corresponding parameters are
$\alpha = r + aW$ and $\beta = s + (1-a)W$, and expectation value $E[x]=\alpha/(\alpha+\beta)=(r+aW)/(r+s+W)$.

\paragraph{Confidence representation} SL separates confidence into two complementary notions which is advantageous for representing confidence in assurance arguments in a nuanced way. \textit{First-order confidence} is the degree of commitment to belief or disbelief and increases as uncertainty $u$ decreases, for example when consistent evidence accumulates or independent sources are fused. \textit{Second-order confidence} is the precision of that commitment: each opinion induces a Beta distribution, with a narrow, peaked shape indicating high precision and a wide, flat shape indicating low precision; precision grows with independent observations and is influenced by the prior weight $W$. This distinction captures both how committed an assessment is and how precisely that assessment can be trusted.

\paragraph{Combining opinions} SL provides a wide range of combination operators \cite{josang2016subjective}. Combining opinions provides an elegant and intuitive way to combine the underlying Beta distributions, a direct manipulation of which would be significantly more complex. In this paper, we use the following operators (see \cite{josang2016subjective} for full definitions):

\textit{\textbf{Conjunction and disjunction}} of opinions about independent claims using the SL equivalents of logical AND and OR, respectively. Let $\omega_x = (b_x,d_x,u_x,a_x)$ and $\omega_Y = (b_y,d_y,u_y,a_y)$ be two independent opinions about two domains of interest $\mathbb{X} = \{x,\bar{x}\}$ and $\mathbb{Y} = \{y,\bar{y}\}$. Opinion $\omega_{x \land y}$ on the conjunction $x \land y$ is computed as $\omega_{x \land y} = \omega_x \cdot \omega_y$ using Binomial multiplication. Likewise, opinion $\omega_{x \lor y}$ on the disjunction $x \lor y$ is computed as $\omega_{x \lor y} = \omega_x \sqcup \omega_y$ using Binomial co-multiplication. 

\textit{\textbf{Cumulative fusion}} of independent opinions about alternative pieces of support for a single claim. This operator represents situations where uncertainty decreases as more independent support is added. Suppose two independent sources $A$ and $B$ formulate opinions over the same binary domain $\mathbb{X}=\{x,\bar{x}\}$, denoted by $\omega^A_x=(b^A_x,d^A_x,u^A_x,a^A_x)$ and $\omega^B_x=(b^B_x,d^B_x,u^B_x,a^B_x)$. The cumulative fusion of these opinions is denoted by $\omega^{A \diamond B}_x = \omega^A_x \oplus \omega^B_x$ (for readability, we write $\omega_x$ instead of $\omega^A_x$ or $\omega^B_x$ when the sources are clear from context). 

\textit{\textbf{Conditional deduction}} of a conclusion opinion from an opinion about a premise under conditional opinions. We assume here that the conclusion depends on the premise through a directional conditional relationship, expressed by the conditional opinions. Let $\mathbb{X}=\{x,\bar{x}\}$ and $\mathbb{Y}=\{y,\bar{y}\}$ be two binary domains. Let $\omega_x = (b_x, d_x, u_x, a_x)$, $\omega_{y|x} = (b_{y|x}, d_{y|x}, u_{y|x}, a_{y|x})$ and $\omega_{y|\bar{x}} = (b_{y|\bar{x}}, d_{y|\bar{x}}, u_{y|\bar{x}}, a_{y|\bar{x}})$ be opinions about $x$ being true, $y$ being true given $x$ is true, and $y$ being true given $x$ is false. Based on that, $\omega_{y||x} = \omega_x \deduction (\omega_{y|x}, \omega_{y|\bar{x}})$ denotes a conditionally deduced opinion. For readability, we abbreviate $\omega_{y||x}$ as $\omega_y$ only when $x$ is clear from context and no additional premises or contexts remain.

\section{Mapping argument relations to SL}
\label{sec:mapping}

We now use the SL operators introduced above to endow each argument relation with a quantitative interpretation to enable systematic propagation of confidence across the argument graph. By assigning SL-based opinions, i.e.\ degrees of belief, disbelief, and uncertainty, to goals and evidential artifacts, and operators to edges (e.g., fusion, conjunction/disjunction, and conditional deduction) we can:
(i) model reasoning strategies and their dependence assumptions;
(ii) make the influence of context explicit and scoped;
(iii) quantify uncertainty and propagate confidence consistently; and
(iv) enable automation for checking, calculation, and what-if analysis.

We assign binomial opinions to the core elements of an assurance argument, i.e., goals, evidence, and assumptions -- as these represent propositions with evaluable truth values and thus support confidence propagation. We further interpret each edge as an operator in SL. The mapping considers two basic relations used in assurance arguments -- \textit{support} and \textit{context}. 

\subsection{Support relations}

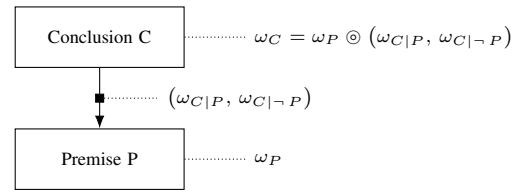
\begin{figure}[b]
\centering
\begin{tikzpicture}[baseline=0, node distance=5mm, every node/.style={font=\scriptsize}]
  \node[gsn-goal, text width=2cm] (G) {Conclusion C};
  \node[gsn-goal, text width=2cm, below=8mm of G] (P) {Premise P};
  \draw[edgeStyle] (G) -- node[midway, fill=black, draw=black, shape=rectangle, inner sep=0pt, minimum size=3pt](acp) {} (P);
  \node[right=7mm of acp] (cond) {$\conditionals{C}{P}$};
  \draw[densely dotted] (acp) -- (cond);

  \node[right=8mm of P] (ox) {$\opinion{P}$};
  \draw[densely dotted] (ox) -- (P);
  
  \node[right=8mm of G, align=left] (oy) {$\opinion{C} = \opinion{P} \deduction \conditionals{C}{P}$};
  \draw[densely dotted] (oy) -- (G);
  
\end{tikzpicture}
\caption{Support relation modelled as conditional deduction. Conclusion claims may be represented as GSN goals, while premises may be evidences or sub-goals.}\label{fig:supportedBy}
\end{figure}

A support relation represents an inferential step where a premise $P$ is linked to a conclusion $C$. In GSN, this is modelled using a \emph{supportedBy} link. We represent this inferential step in SL as a conditional deduction from premise to conclusion, using conditional opinions that encode the required justification. Let $P$ denote the premise representing a goal or evidence, and let $C$ be the conclusion element representing a goal. The support relation is then modelled as follows:
\begin{align}\label{eq:supportedBy}
    \omega_C &= \omega_P \deduction \conditionals{C}{P}
\end{align}

where $\conditionals{C}{P}$ are the conditional opinions capturing the required justification for the inference. For the remainder of this paper, for readability, we abbreviate $\omega_{C||P}$ as $\omega_C$ because $P$ is clear from the context here and no additional premises or contexts remain to be applied. A graphical description of the mapping is given in Fig.~\ref{fig:supportedBy}. Premise $P$ is assigned a Binomial opinion $\omega_P$. As described in more detail in Section \ref{sec:justifying}, conditional opinions $\conditionals{C}{P}$ serve as the justification for the link between $P$ and $C$ (as denoted by the black square on the edge). The target opinion $\omega_C$ is then formally deduced from the premise opinion $\omega_P$ using the justification. 

\subsection{Context relations}\label{sec:inContextOf}

Assurance elements with a contextual relationship, such as assumptions, justifications, or context nodes, can be attached to claims or strategies. GSN provides the \emph{inContextOf} link for this purpose. We focus here on assumptions, which are particularly relevant \textit{because they condition the element that they are attached to}. In other words, the validity of the element (such as a goal) depends on the validity of the assumptions. Consider, for example, a goal $G$ with an attached assumption $A$ via the \emph{inContextOf} relationship. Intuitively, the truth of target $G$ is only asserted under $A$. The relation between assumption $A$ and goal $G$ can thus be modelled as an explicit condition on $G$. To make this visible within the argument, we treat the goal $G$ as a \textit{conditional claim} $G \given A$ (instead of just $G$) and attach the conditional opinion $\opinion{G \given A}$. It is now clearly obvious that all supporting elements beneath $G$ (sub-goals, evidence) are interpreted as \textit{justifying $G$ under $A$}, so confidence is composed locally within the stated context.

There are two possible ways to treat the contextual elements, depending on whether (1) the conditional opinion about the element within its context should be kept, or (2) the element should be combined with its context in order to obtain an unconditional opinion about the element. 

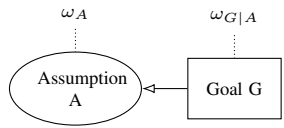
\begin{figure}[b]
\centering
\begin{tikzpicture}[baseline=0, node distance=6mm, every node/.style={font=\scriptsize}]
  \node[gsn-goal] (G) {Goal G};
  \node[above=3mm of G] (oG) {$\opinion{G \given A}$};
  \draw[densely dotted] (oG) -- (G);
  \node[gsn-assumption, left=of G] (A) {Assumption A};
  \node[above=3mm of A] (oA) {$\opinion{A}$};
  \draw[densely dotted] (oA) -- (A);
  \draw[-{Latex[open]}] (G.west) -- (A);
\end{tikzpicture}
\caption{Context  relation: conditional interpretation}
\label{fig:context_conditional}
\end{figure}

\paragraph{Conditional interpretation} under the conditional interpretation, we simply keep $\opinion{G \mid A}$ and propagate the conditional claim to parents without marginalising. When propagated to parent goals, the context is thus also propagated: parent claims remain conditional on $A$ (and accumulate further contexts similarly). This keeps assumptions explicit and prevents hidden mixing of contexts. A graphical description of this mapping is shown in Fig.~\ref{fig:context_conditional}.

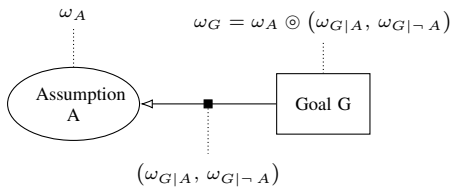
\begin{figure}[b]
\centering
\begin{tikzpicture}[baseline=0, node distance=8mm, every node/.style={font=\scriptsize}]
  \node[gsn-goal] (G) {Goal G};
  \node[gsn-assumption, left=18mm of G] (A) {Assumption A};

  \draw[-{Latex[open]}] (G.west) -- node[midway, fill=black, draw=black, shape=rectangle, inner sep=0pt, minimum size=3pt] (acp) {} (A);
  \node[below=6mm of acp] (cond) {$\conditionals{G}{A}$};
  \draw[densely dotted] (acp) -- (cond);
  

  \node[above=4mm of G] (oG) {$\opinion{G} = \opinion{A} \deduction \conditionals{G}{A}$};
  \draw[densely dotted] (oG) -- (G);

  \node[above=5mm of A] (oA) {$\opinion{A}$};
  \draw[densely dotted] (oA) -- (A);
\end{tikzpicture}
\caption{Context relation: unconditional interpretation}
\label{fig:context_unconditional}
\end{figure}

\paragraph{Unconditional interpretation} should an unconditional opinion be required (e.g., for overall confidence assessment), it can be produced by \textit{marginalising} the context, i.e.\ by ``consuming'' $A$ through conditional deduction. Technically, assumption $A$ becomes the premise that corresponding goal $G$ is deduced from, and conditional opinion $\omega_{G|A}$ becomes one of the two conditionals needed to justify the link between assumption $A$ and goal $G$: 
\begin{align} 
\opinion{G} = \opinion{A} \deduction \conditionals{G}{A}
\end{align}

Provenance is clearly preserved: Assumption $A$ is marked as consumed at goal $G$ and should not be reapplied upstream. A graphical depiction of this interpretation is shown in Fig.~\ref{fig:context_unconditional}. The \emph{inContextOf} edge from $G$ to $A$ carries conditionals $\conditionals{G}{A}$ as the justification (denoted by the black square in Figure \ref{fig:context_unconditional}). The goal $G \given A$ remains visible with its own opinion and represents the first (positive) conditional $\opinion{G \given A}$. The unconditional opinion $\omega_G$ is then computed by marginalising the context: $\opinion{G} = \opinion{A} \deduction \conditionals{G}{A}$. Context $A$ is consumed at $G$ to prevent reapplication upstream.

This design has two practical benefits. First, it offers a uniform, operator-based semantics (everything is based on conditional deduction) and avoids special-case handling of context. Second, it separates modelling from presentation: users work with explicit conditional claims ($G \mid context$), and marginalisation is applied only where an unconditional view is required, with safeguards to prevent double discounting and to maintain auditability.

\section{Modular composition of safety arguments}
\label{sec:argument_patterns}

We demonstrate now how more complex arguments and their associated SL confidence networks can be assembled in a modular and systematic manner, enhancing both the expressiveness and compositionality of assurance arguments. Whilst the approach is notation-agnostic, we use the GSN argument shown in Fig.~\ref{fig:gsn-running-example} as a running example for illustration. We decompose the argument into \textit{argument patterns} \cite{YuanSubjectiveLogic2017, Duan2015}, i.e., composite building blocks, and show how they can be constructed. We start with the sub-argument comprising goal \textit{G4} and evidence \textit{Sn1} which represents a \textit{one-to-one} argument. 

\subsection{One-to-One relations}

A one-to-one argument represents the scenario where a single conclusion element $C$ is supported by a premise element $P$. In the argument in Fig.~\ref{fig:gsn-running-example}, solution $Sn1$ provides direct support for goal $G4$. This represents a single support relation as described in Fig.~\ref{fig:supportedBy}. The satisfaction of $G4$ depends on the validity of $Sn1$, and our opinion in $G4$ is deduced from our opinion in $Sn1$ using the conditional deduction operator as described in Eq.\ \eqref{eq:supportedBy} as follows: 

\begin{equation} \nonumber
\opinion{G4} = \opinion{Sn1} \deduction \conditionals{G4}{Sn1}\label{eq:one-to-one}
\end{equation}
A one-to-one argument thus maps directly to a conditional deduction in SL. In this case, two conditional opinions are required to compute $\omega_{G4}$. They represent the validity of the inferential step between $\omega_{Sn1}$ and $\omega_{G4}$. In GSN, this could be explicated by a strategy between $G4$ and $Sn1$ with an optional justification that explains why the inference holds, by assurance claim points (ACPs) \cite{hawkins2010new}, or in a dialectical way using defeaters \cite{herd2025integrating}. We return to this point in Section \ref{sec:justifying}. 

\subsection{One-to-Many relations}
One-to-many arguments represent scenarios where a single conclusion $C$ is supported by multiple premise elements $P_1 \cdots P_n$. In GSN, strategy nodes serve as descriptors of argument relations. Conceptually, an argumentative strategy is always present when a one-to-many relation is defined, whether this is made explicit through a strategy node or remains implicit. In this work, we consider such argumentative strategies and propose to map them to specific categories, following the reasoning semantics of assurance arguments. To this end, we build upon \cite{Duan2015,YuanSubjectiveLogic2017} and distinguish between two categories of one-to-many relations, classified according to the domain to which the claims refer: \textit{fusion} (combination of claims within the same domain) and \textit{logical combination} (combination of claims across different domains). We describe the two categories below. 

\subsubsection{Fusion}
Fusion arguments describe situations where several sources of evidence inform an opinion about the same claim. Their purpose is to aggregate these sources in a way that faithfully reflects the intended semantics of the argument. SL provides dedicated fusion operators for these cases, as described in \cite{josang2016subjective}. For this paper, we focus on a subset of them: (1) \textit{cumulative fusion}, used when evidence sources are independent; (2) \textit{averaging fusion}, applied when evidence is dependent; and (3) \textit{weighted fusion}, which accounts for varying confidence in the evidence.

The sub-argument that involves goals $G2$, $G4$, and $G5$, connected through strategy $S2$ (labelled II in Fig.~\ref{fig:gsn-running-example}) represents an example of a fusion argument; suppose sources $A$ and $B$ formulate opinions on $G4$ and $G5$, respectively, each providing independent support for $G2$'s claim that hazard 1 has been mitigated. This relationship can be modeled by (1) aggregating $G4$ and $G5$ using a fusion operator and (2) connecting the resulting opinion with $G2$ using a support relationship: 
\begin{align} 
\omega^{A \diamond B} &= \opinion{G4} \oplus \opinion{G5}, \\ 
\opinion{G2} &= \opinion{G4} \oplus \opinion{G5} 
\deduction \conditionals*{G2}{\opinion{G4} \oplus \opinion{G5}}
\end{align}
In this case, we assume independence of the evidence and use the cumulative fusion ($\oplus$) operator.    

\subsubsection{Logical combination}
The second category of one-to-many relations concerns argument structures in which independent opinions over \textit{distinct} domains are logically combined using operations such as conjunction (AND) and disjunction (OR) of independent branches of reasoning. For example, demonstrating system safety may require that both functional correctness and fault-tolerance claims hold simultaneously (conjunction), or may alternatively be supported by evidence of effective fallback mechanisms (disjunction).   

\paragraph{Conjunctive argument}\label{sec:conjunctive_argument}
In a conjunctive argument, truth of all premises is required to support the conclusion. If any premise is false, the conclusion does not hold. This corresponds to a logical AND relation which can be represented in SL using the Binomial multiplication operator ($\cdot$), followed by a single inferential support relationship. The sub-argument where goal $G1$ is supported by goals $G2$ and $G3$ (labelled I in Fig.~\ref{fig:gsn-running-example}) is conjunctive. According to strategy $S1$, both $G2$ and $G3$ need to be satisfied in order for $G1$ to also be satisfied. This can be modelled by (1) aggregating $G2$ and $G3$ using Binomial multiplication, and (2) connecting the resulting opinion with $G1$ using an inferential support relation: 
\begin{align} 
\opinion{G2 \wedge G3} &= \opinion{G2} \cdot \opinion{G3} \\ 
\opinion{G1} &= \opinion{G2 \wedge G3} 
\deduction \conditionals*{G1}{G2 \wedge G3}
\end{align}

As described in Section \ref{sec:inContextOf}, assumption $A1$ attached to $G1$ can be dealt with by either viewing $G1$ as a conditional claim or by `consuming' $A1$ and computing a marginalized opinion $\omega_{G1}$ through conditional deduction. 

\paragraph{Disjunctive argument}
In a disjunctive argument, any one of the premises can support the conclusion. This corresponds to a logical OR relation, represented in SL using the co-multiplication operator ($\sqcup$), followed by a single inferential support relationship. Each premise provides an alternative path to satisfy the goal. The sub-argument comprising goals $G3$, $G6$, and $G7$, connected with strategy $S3$ (labelled III in Fig.~\ref{fig:gsn-running-example}) is disjunctive. As stated by $S3$, $G6$ and $G7$ refer to alternative mitigation measures, each of which is able to satisfy $G3$ independently. This can be modelled by (1) aggregating $G6$ and $G7$ using Binomial co-multiplication, and (2) connecting the resulting opinion with $G3$ using a support relation:
\begin{align} 
\opinion{G6 \vee G7} &= \opinion{G6} \sqcup \opinion{G7} \\ 
\opinion{G3} &= \opinion{G6 \vee G7} 
\deduction \conditionals*{G3}{G6 \vee G7}
\end{align}

\subsection{Justifying inferential steps}\label{sec:justifying}

Every support and context relation asserts an implicit warrant that a premise renders its conclusion credible, or that a context condition defines the scope within which a claim is to be read. In our semantics, each such edge is justified explicitly by a pair of conditional opinions that parameterize deduction. For inferential support from premise $P$ to conclusion $C$, the interface is the ordered pair $\conditionals{C}{P}$ used in 
\begin{align}
\omega_{C} = \omega_{P} \deduction \conditionals{C}{P}
\end{align}
where $\deduction$ denotes the conditional deduction operator introduced in Section \ref{sec:subjective_logic}. For contextual attachment of assumption $A$ to goal $G$ under the unconditional interpretation (see Fig.~\ref{fig:context_unconditional}), the corresponding pair $\conditionals{G}{A}$ serves to marginalise $A$ into an unconditional $\omega_{G}$:
\begin{align}
\omega_G &= \omega_A \deduction \conditionals{G}{A}
\end{align}
Thus, edges are quantitative interfaces whose conditionals constitute the formal warrant for link validity.

Practitioners can realize these warrants in multiple, complementary ways in practice. In GSN, a common pattern is to attach an informal justification node that explains the rationale, applicability conditions, and limits of the inference or context; these justifications are then translated into the edge conditionals using the elicitation methods described below. When arguments cross organisational or architectural boundaries, a modular interface can be defined whose deliverable are the conditional opinions required by the link. Assurance Claim Points (ACPs) \cite{hawkins2010new} could be one such interface mechanism, but our semantics does not depend on ACPs per se; any interface that yields $\conditionals{G}{P}$ or $\conditionals{G}{A}$ integrates identically via conditional deduction.

\section{Overall Confidence Assessment}\label{sec:example}
Based on the building blocks described in the previous section, we demonstrate the overall confidence assessment on the simple assurance argument presented in Fig.~\ref{fig:gsn-running-example}. We conduct the assessment by following four steps: 

\begin{enumerate}
    \item \textbf{Identify argument reasoning strategies.} We first identify and map argument reasoning strategies, either explicitly stated or inferred, to the argument types in Section~\ref{sec:argument_patterns}. 
    \item \textbf{Elicit opinions.} We define conditional opinions along edges and assign opinions to input nodes. 
    \item \textbf{Propagate confidence.} We propagate confidence through the argument using SL operators. 
    \item \textbf{Review and iterate.} We address context relations in the argument, review the overall confidence, revise the argument, and repeat as needed until satisfactory confidence is reached.
\end{enumerate}

We now proceed systematically through these steps. 

\paragraph{Identification of argument reasoning strategies}
The first step is to identify and characterize the reasoning strategies used in the argument, based on their underlying \emph{purpose}. Each strategy defines how supporting claims contribute to their parent goal. Strategies that aim to \emph{divide and conquer} a complex claim are modeled as conjunctive arguments, requiring all sub-claims to hold simultaneously. Those that \emph{provide alternative argumentation paths} are treated as disjunctive arguments, where any one sub-claim is sufficient to support the conclusion. Finally, strategies that \emph{strengthen the argument through redundancy} are represented as fusion arguments: (1) \emph{cumulative fusion} when the supporting evidence is independent, (2) \emph{averaging fusion} for dependent evidence, and (3) \emph{weighted fusion} when differences in evidence confidence should be accounted for.

In Fig.~\ref{fig:gsn-running-example}, three strategy nodes are explicitly defined, each representing a distinct argument type: (I) conjunctive, (II) fusion, and (III) disjunctive, as summarized in Table~\ref{tab:strategy-mapping}.

\begin{table}[b]
\caption{Mapping of strategies to argument types in SL.}
\centering
\begin{tabularx}{\linewidth}{
  >{\raggedright\arraybackslash}p{0.08\linewidth}
  >{\raggedright\arraybackslash}p{0.3\linewidth}
  >{\centering\arraybackslash}p{0.26\linewidth}
  >{\centering\arraybackslash}X
}
\toprule
\textbf{Strategy} & \textbf{Purpose} & \textbf{Argument type} & \textbf{Formulation in SL} \\
\midrule
S1 & Divide and conquer & Conjunction  & $\omega_{G2} \cdot \omega_{G3}$  \\
S2 & Provide alternative argumentation paths & Disjunction  & $\omega_{G6} \sqcup \omega_{G7}$ \\
S3 & Strengthen argument through redundancy & Cumulative fusion & $\omega_{G4} \oplus \omega_{G5}$ \\
\bottomrule
\end{tabularx}
\label{tab:strategy-mapping}
\end{table}

\paragraph{Opinion elicitation}
Before propagating confidence, we must first elicit the required conditional opinions, as well as the input opinions. Eliciting edge conditionals should be systematic and reviewable. When data are available, evidence-based elicitation counts positive and negative instances for the conditional claim, sets a base rate $a$, and converts to a binomial opinion using the standard mapping in Eqs.~\eqref{eq:belief}--\eqref{eq:uncertainty} with prior weight $W$ (commonly $W=2$). If data for the negative conditional (e.g., $\omega_{C\mid \neg P}$ or $\omega_{G\mid \neg A}$) are sparse, a conservative prior with high disbelief should be used and documented. Where empirical counts are unavailable, expert judgment can be elicited using calibrated qualitative categories that combine likelihood and confidence, with a transparent mapping to $(b,d,u,a)$ \cite{pope2005analysis}. In both cases, the resulting conditionals form the edge-level quantitative warrant and can, when appropriate, be produced by a focused confidence sub-argument whose endpoint is the conditional of interest. The focus of this paper is not on the elicitation of the opinions themselves.
For all conclusions $C$ and premises $P$ in Fig.~\ref{fig:gsn-running-example} that are connected by a support relation, we define the positive conditional opinions to have high belief and low uncertainty, namely $\opinion{C \given P}$ = $(b=0.95, d=0.00, u=0.05)$. To reflect a conservative reasoning stance, where the premises are considered necessary for the conclusion, we set the corresponding negative conditionals $\opinion{C \given \neg P}$ to full disbelief ($b=0, d=1, u=0$). 

\begin{figure}[b]
    \centering
    \includegraphics[width=0.7\columnwidth]{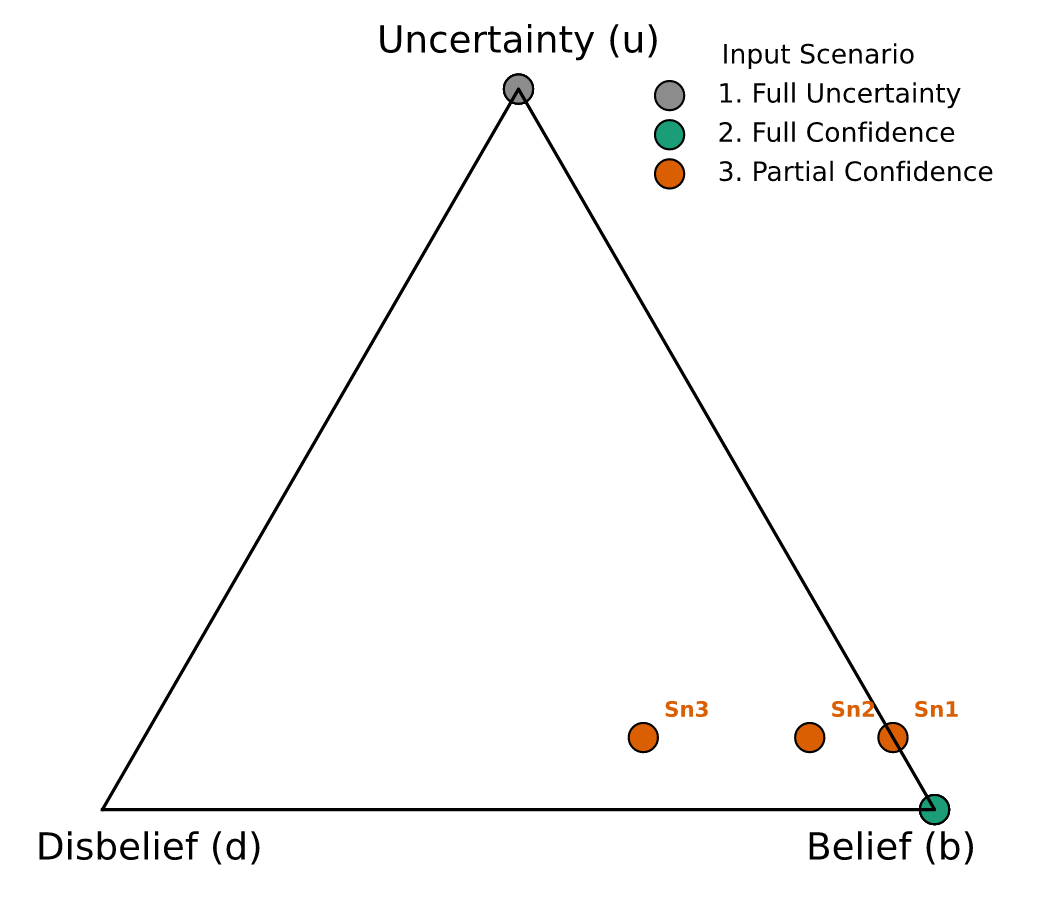}
    \caption{
        Input opinion configurations for the evidence nodes (\textit{Sn1–Sn3}) in the argument model.
        Each vertex represents belief ($b$), disbelief ($d$), and uncertainty ($u$).
        Three input scenarios are shown: full uncertainty, full confidence, and partial confidence.
        Labels indicate the individual evidence opinions under the \emph{Partial Confidence} scenario.
    }
    \label{fig:input-opinions-triangle}
\end{figure}

We fix the conditional opinions and consider three \textbf{input opinion scenarios} for the evidence nodes (\textit{Sn1–Sn3}) in Fig.~\ref{fig:gsn-running-example}. These three configurations are visualized using an opinion triangle \cite{josang2016subjective} in Fig.~\ref{fig:input-opinions-triangle}. In the full uncertainty scenario, all input nodes are vacuous ($u=1$). In the full confidence scenario, each input node expresses total belief $(b=1)$. The partial confidence scenario introduces more realistic variability, with residual uncertainty across the inputs: $\opinion{Sn1}=(0.9, 0.0, 0.1)$, $\opinion{Sn2}=(0.8, 0.1, 0.1)$, and $\opinion{Sn3}=(0.6, 0.3, 0.1)$.

\paragraph{Propagation}
Given the input node opinions, we propagate the confidence along the edges using the relevant operators. The resulting opinions are presented in Table~\ref{tab:opinion-high}. In the full uncertainty scenario, the evidence nodes contribute no belief, and complete uncertainty is propagated to the top level. In the full confidence scenario, the resulting top-level opinion exhibits high belief and low uncertainty, with the remaining uncertainty arising from the deduction conditionals that mediate the support relations. In the partial confidence scenario, the outcome reflects moderate levels of belief, disbelief, and uncertainty, corresponding to the mixed strength of the input evidence.

\begin{figure}[b]
    \centering
    \includegraphics[width=0.9\columnwidth]{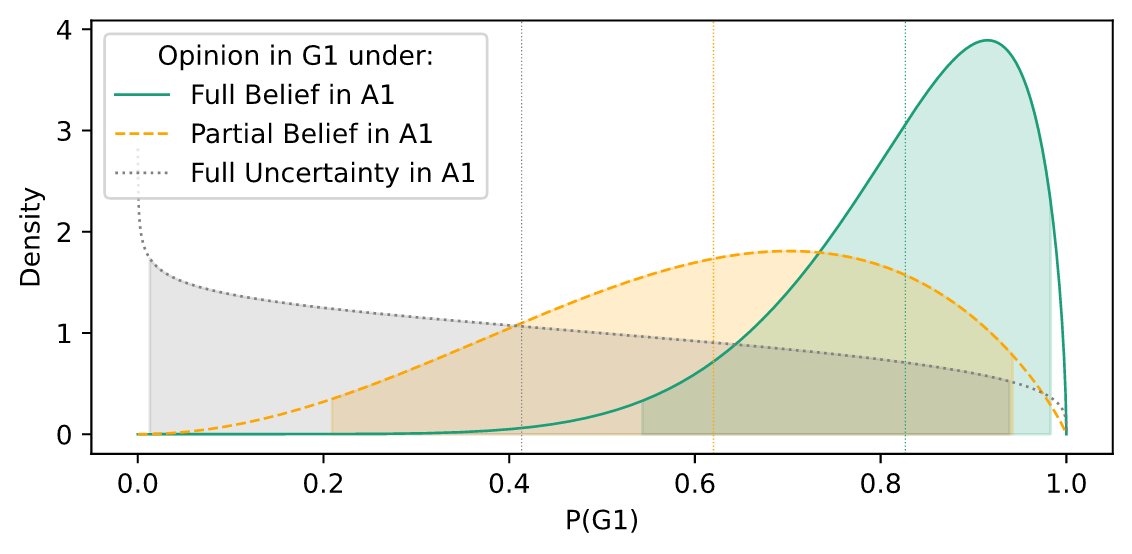}
    \caption{Top-level opinion distributions ($\opinion{G1}$) under three assumption opinions ($\opinion{A1}$): 
    full uncertainty, partial belief, and full belief. Each curve shows the probability $P(G1)$ after considering $\opinion{A1}$ under the \emph{High Belief} conditional scenario 
    with \emph{Partial Confidence} input opinions. The dotted vertical lines mark the expected probability for each case.}
    \label{fig:g1-beta-absorption}
\end{figure}

\paragraph{Review and iterate}
To assess the top level confidence, we first analyze how assumptions influence these propagation results by varying the opinion of $\opinion{A1}$ and observing its impact on top-level opinion $\opinion{G1}$ (see Fig.~\ref{fig:g1-beta-absorption}). Specifically, we vary $\opinion{A1}$ and recompute $\opinion{G1}$ by conditioning on $\opinion{A1}$.  We consider three assumption opinions: a \emph{Full Belief} case $(b=1.0, d=0.0, u=0.0, a=0.5)$ representing complete confidence in the assumption, a \emph{Partial Belief} case $(b=0.6, d=0.3, u=0.1, a=0.5)$ representing moderate support, and a \emph{Full Uncertainty} case $(b=0.0, d=1.0, u=0.0, a=0.5)$ representing total disbelief.
The resulting beta distributions for the top-level opinions show how confidence in $A1$ influences $\opinion{G1}$. Fig.~\ref{fig:g1-beta-absorption} shows that decreasing confidence in $A1$ reduces both first-order and second-order confidence in $G1$. We can see the overall opinion in $\opinion{G1}$ in the top row of Table \ref{tab:opinion-high}. 

\begin{table*}[!t]
\caption{Opinion propagation for the argument in Figure~\ref{fig:gsn-running-example}. The table shows results for the three input scenarios, illustrating how the top-level opinion is composed from the evidence node opinions through successive applications of the relevant Subjective Logic operators.}
\label{tab:opinion-high}
\begin{tabular}{@{}lllll@{}}
\toprule
Node & Opinion Computation & Full Uncertainty Inputs & Full Confidence Inputs & Partial Confidence Inputs \\
\midrule
 \multicolumn{5}{c}{\emph{Top-Level Goal}} \\[0.2em] \midrule \\
$G1$ & $\opinion{G1} = \opinion{A1} \deduction \conditionals{A1}{(G1 \given A1)}$ & \textbf{(0.00, 0.00, 1.00)} & \textbf{(0.86, 0.00, 0.14)} & \textbf{(0.74, 0.05, 0.21)} \\ 
$G1 \given A1$ & $\opinion{G1 \given A1} = \opinion{G2 \wedge G3} \deduction \conditionals{G1}{(G2 \wedge G3)}$ & \textbf{(0.00, 0.00, 1.00)} & \textbf{(0.86, 0.00, 0.14)} & \textbf{(0.74, 0.05, 0.21)} \\ \midrule
 \multicolumn{5}{c}{\emph{Intermediate Goals}} \\[0.2em] \midrule \\
$G2$ & $\opinion{G2} = \opinion{G4 \oplus G5} \deduction \conditionals{G2}{(G4 \oplus G5)}$ & (0.00, 0.00, 1.00) & (0.91, 0.00, 0.09) & (0.82, 0.00, 0.18) \\
$G3$ & $\opinion{G3} = \opinion{G6 \vee G7} \deduction \conditionals{G3}{(G6 \vee G7)}$ & (0.00, 0.00, 1.00) & (0.95, 0.00, 0.05) & (0.85, 0.05, 0.10) \\
$G4$ & $\opinion{G4} = \opinion{Sn1} \deduction \conditionals{G4}{Sn1}$ & (0.00, 0.00, 1.00) & (0.95, 0.00, 0.05) & (0.86, 0.00, 0.14) \\
$G5$ & Full uncertainty default opinion. & (0.00, 0.00, 1.00) & (0.00, 0.00, 1.00) & (0.00, 0.00, 1.00) \\
$G6$ & $\opinion{G6} = \opinion{Sn2} \deduction \conditionals{G6}{Sn2}$ & (0.00, 0.00, 1.00) & (0.95, 0.00, 0.05) & (0.76, 0.10, 0.14) \\
$G7$ & $\opinion{G7} = \opinion{Sn3} \deduction \conditionals{G7}{Sn3}$ & (0.00, 0.00, 1.00) & (0.95, 0.00, 0.05) & (0.57, 0.30, 0.13) \\
\bottomrule
\end{tabular}
\end{table*}

\section{Related Work}\label{sec:related_work}

A substantial body of work has explored formal or compositional semantics for assurance arguments \cite{bandur2014informing,graydon2015formal,denney2013formal,denney2014towards,denney2016composition,hawkins2015weaving}, focusing primarily on logical soundness, structural pattern reuse, or traceability. In contrast, our work introduces a semantics explicitly designed for quantitative confidence assessment, where SL serves as the underlying calculus of uncertainty propagation and representation. We thus restrict the focus to related work in that area.  

A general quantification of assurance cases is provided by Duan et al. \cite{Duan2015}, using the beta distribution representation of SL. The consensus and logical-OR operator illustrate how combining different beliefs from evidence to the main claim can be performed. By that, they show the general applicability of SL in assurance cases, while concluding that real assurance cases might need more complex and unique operators than the atomic ones that are part of SL. As a step towards narrowing this gap, we leverage additional SL operators to reflect a broader range of argument reasoning structures.

Yuan et al. \cite{YuanSubjectiveLogic2017} extend \cite{Duan2015} by introducing four operator templates for one-to-one, conjunctive, disjunctive, and alternative arguments. While these are composed of the same components as our approach (deduction, co-/multiplication, fusion), the order of operation differs. Alternative and disjunctive arguments decompose the parent claim into subclaims represented by a single premise. Each of these premises' beliefs are individually decomposed into subclaims, which are then combined with the co-multiplication or fusion operator. This relies on the assumption of no additional uncertainty being introduced by the decomposition strategy itself, which is what our model is able to include. The conjunctive argument uses deductions both before and after the binomial multiplication. We use a single compositional step that mirrors the argument's actual structure.

Idmessaoud et al. \cite{IdmessaoudConfidenceAssessment2024} present a different approach for a confidence propagation model, based on propositional logic expressions and quantification of uncertainty with Dempster-Shafer theory. They also distinguish between uncertainty in the premise itself, and uncertainty in the support relation between a premise and its conclusion. From that, they build four argument types: simple (one-to-one), conjunctive, disjunctive, and hybrid arguments, matching the templates of \cite{YuanSubjectiveLogic2017} and our approach. All three of their many-to-one arguments consist of a combination and a deduction step, but include both the single premise-to-conclusion uncertainty, as well as the combined-premises-to-conclusion uncertainty. Since it is more effort to derive the belief values for all of these inferences, they also propose a procedure of confidence assessment. Unlike their dual quantitative and qualitative approach, we employ SL to unify both kinds of uncertainty in a single probabilistic model. 

\section{Discussion and conclusions}\label{sec:discussion}

This paper introduces a compositional semantics for quantitative confidence assessment in assurance arguments which enables the formal representation of end-to-end propagation of belief, disbelief, and uncertainty using Subjective Logic (SL). By interpreting argument elements as SL opinions and every relation as a corresponding operator, the approach provides a unified framework for quantifying confidence across assurance arguments. This enables users to quantify and trace how confidence in top-level claims arises from underlying evidence, identify weak links in the argument, and systematically explore the effect of new or revised evidence. From a continuous development perspective, this work complements recent work on change-impact analysis \cite{carlan2022automating} which proposes a semi-automated method to trace and quantify how changes in ML components affect related artifacts such as datasets of performance requirements. Our approach assesses how confident we are in the \textit{current evidence} and its composition. The ability to re-evaluate confidence after each iteration offers a direct mechanism for assessing the effect of change through updated confidence. This separation of change analysis and confidence assessment supports frequent re-assessment during iterative AI system development. 

Several open challenges remain to be addressed in future work. At the current stage, the proposed semantics focusses on \textit{support} and \textit{context} relations. Future work should further extend the semantics, e.g.\ to achieve full compability with the Structured Assurance Case Metamodel (SACM) \cite{omg_sacm_2020}. Furthermore, logical combination and cumulative fusion operators in SL rely on the assumption of evidential independence. In practice, this assumption is often violated -- e.g., in cases where multiple metrics rely on the same dataset or AI model. A systematic treatment of evidential dependence, possibly by representing shared information sources, is an important next step. Another practical challenge concerns the elicitation of the conditional opinions. The number of required conditionals increases with argument complexity, which calls for structured elicitation procedures. Moreover, practitioners require intuitive mappings between SL opinions and qualitative confidence levels to facilitate communication with stakeholders. 

While existing work shows that quantitative confidence assessment has a strong theoretical base, its practical adoption is still weak. According to a recent study  \cite{DIEMERT2025107767}, the main concern is the increased effort  combined with inadequate methodological guidance in creating and interpreting quantitative methods. This signifies a gap between theoretical research and practitioners' needs. The ability to communicate a quantitative confidence assessment method in an intuitive and standardized way will thus be a key quality in deciding its success. While our current approach still requires significant involvement from a human expert, future work may include automated tool support. Scalability and automation will ultimately determine the practical applicability of the approach. Automating the transformation from notations like GSN into SL networks would facilitate `what-if' analyses and continuous confidence monitoring, and help to close the gap between qualitative reasoning and quantitative assurance. 

\section{Acknowledgments}

The research leading to these results is funded by the German Federal Ministry for Economic Affairs and Energy within the project ``Safe AI Engineering -- Sicherheitsargumentation befähigendes AI Engineering über den gesamten Lebenszyklus einer KI-Funktion''. The authors would like to thank the consortium for the successful cooperation.

\bibliographystyle{ieeetr}
\bibliography{references}
\end{document}